\documentclass[conference]{IEEEtran}
\IEEEoverridecommandlockouts

\usepackage{cite}
\usepackage{amsmath,amssymb,amsfonts}
\usepackage{algorithmic}
\usepackage{graphicx}
\usepackage{textcomp}
\usepackage{xcolor}
\usepackage{hyperref}
\usepackage{multirow}
\usepackage{arydshln}
\def\BibTeX{{\rm B\kern-.05em{\sc i\kern-.025em b}\kern-.08em
    T\kern-.1667em\lower.7ex\hbox{E}\kern-.125emX}}
\begin{document}

\title{Deformable Dynamic Convolution for Accurate yet Efficient Spatio-Temporal Traffic Prediction}

\author{\IEEEauthorblockN{1\textsuperscript{st} Hyeonseok Jin}
\IEEEauthorblockA{\textit{Dept. of Artificial Intelligence Convergence} \\
\textit{Chonnam National University}\\
Gwangju, South Korea \\
ggyo003@jnu.ac.kr}
\and
\IEEEauthorblockN{2\textsuperscript{nd} Geonmin Kim}
\IEEEauthorblockA{\textit{Dept. of Software Engineering} \\
\textit{Chonnam National University}\\
Gwangju, South Korea \\
204869@jnu.ac.kr}
\and
\IEEEauthorblockN{3\textsuperscript{rd} Kyungbaek Kim\textsuperscript{*}}\thanks{Corresponding Author.}
\IEEEauthorblockA{\textit{Dept. of Artificial Intelligence Convergence} \\
\textit{Chonnam National University}\\
Gwangju, South Korea \\
kyungbaekkim@jnu.ac.kr}
}

\maketitle

\begin{abstract}
Traffic prediction is a critical component of intelligent transportation systems, enabling applications such as congestion mitigation and accident risk prediction.
While recent research has explored both graph-based and grid-based approaches, key limitations remain.
Graph-based methods effectively capture non-Euclidean spatial structures but often incur high computational overhead, limiting their practicality in large-scale systems.
In contrast, grid-based methods, which primarily leverage Convolutional Neural Networks (CNNs), offer greater computational efficiency but struggle to model irregular spatial patterns due to the fixed shape of their filters.
Moreover, both approaches often fail to account for inherent spatio-temporal heterogeneity, as they typically apply a shared set of parameters across diverse regions and time periods.
To address these challenges, we propose the Deformable Dynamic Convolutional Network (DDCN), a novel CNN-based architecture that integrates both deformable and dynamic convolution operations.
The deformable layer introduces learnable offsets to create flexible receptive fields that better align with spatial irregularities, while the dynamic layer generates region-specific filters, allowing the model to adapt to varying spatio-temporal traffic patterns.
By combining these two components, DDCN effectively captures both non-Euclidean spatial structures and spatio-temporal heterogeneity.
Extensive experiments on four real-world traffic datasets demonstrate that DDCN achieves competitive predictive performance while significantly reducing computational costs, underscoring its potential for large-scale and real-time deployment.
\end{abstract}

\begin{IEEEkeywords}
Spatiotemporal prediction, traffic prediction, convolutional neural network, deep learning.
\end{IEEEkeywords}

\section{Introduction}
Traffic prediction plays a critical role in intelligent transportation systems, supporting applications such as dynamic route planning, congestion mitigation, and accident prevention~\cite{xie2020urban,shaygan2022traffic,wang2024citycan}. 
The core task involves forecasting future traffic conditions using historical spatio-temporal data collected from a vast network of sensors deployed across urban regions. 
With the increasing availability of large-scale traffic datasets and the advancement of deep learning, recent studies have shown promising progress in accurately modeling complex traffic dynamics~\cite{zhao2021event, jiang2021dl}.

\begin{figure}[t]
\centering
\includegraphics[width=1.0\columnwidth]{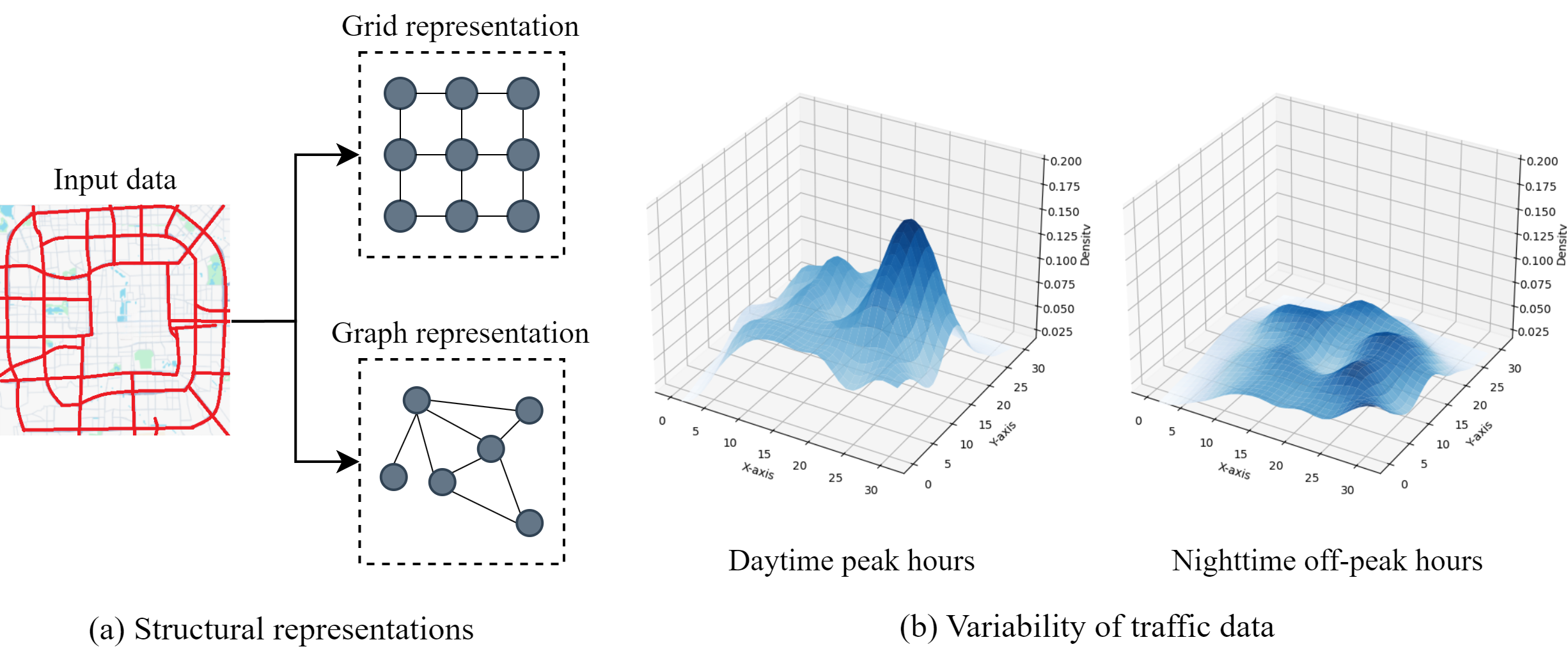}
\caption{An illustration of characteristics of urban traffic data. (a): Different structural representations, (b): spatio-temporal variability.}
\label{fignew}
\end{figure}

As shown in Figure~\ref{fignew} (a), spatio-temporal traffic prediction approaches typically fall into two categories: grid-based and graph-based methods. 
In grid-based methods, CNNs and LSTMs have been widely adopted due to their proven success in computer vision and sequence modeling~\cite{jiang2021dl}. 
However, the inherently non-Euclidean nature of urban road networks where traffic flows are distributed irregularly across heterogeneous regions, has led to the rise of graph-based methods~\cite{peng2024overview}. 
In graph-based methods, Graph Neural Networks (GNNs) offer the advantage of modeling topological dependencies by leveraging graph structures to represent spatial correlations between road segments~\cite{yu2018spatio,diao2023dmstg}.

Despite their success, these approaches face notable limitations. 
GNN-based models, while effective in capturing spatial dependencies, often require pre-defined adjacency matrices and involve computationally expensive graph operations. 
This makes them difficult to scale to high-resolution or city-wide deployments where real-time inference is required~\cite{jin2023spatio,rajagopal2025efficient}. 
Conversely, CNN-based models are computationally efficient and hardware-friendly~\cite{gu2018recent}, benefiting from parallel processing and compatibility with modern accelerators. 
However, traditional CNNs struggle to represent non-Euclidean structures due to their fixed kernel design.
Furthermore, as shown in Figure~\ref{fignew} (b), both approaches often struggle to capture spatio-temporal heterogeneity due to the shared set of parameters.

In real-world deployments such as smart city infrastructures and dynamic traffic signal control, prediction models must balance accuracy with computational efficiency. 
These systems demand rapid inference to support real-time decision-making while handling large volumes of fine-grained traffic data~\cite{rajagopal2025efficient}. 
Therefore, it is essential to design models that offer expressive spatio-temporal representations without sacrificing scalability or responsiveness.

To address these challenges, we propose a novel CNN-based architecture, the Deformable Dynamic Convolutional Network (DDCN). 
DDCN is designed to overcome the inherent limitations of conventional CNNs by incorporating flexible convolutional mechanisms, thereby enabling efficient and expressive modeling tailored for traffic prediction tasks.
Specifically, it integrates two key mechanisms:
(1) a deformable convolution layer that learns spatial offsets to adjust the receptive field according to local irregularities, and
(2) a dynamic convolution layer that generates region-specific filters conditioned on the input, allowing the model to adapt to spatial heterogeneity.

Unlike conventional grid-based methods that typically adopt CNN or ConvLSTM~\cite{shi2015convolutional} backbones, DDCN adopts a Transformer-style CNN backbone with an encoder-decoder structure~\cite{jin2025datc}, which enhances predictive accuracy while preserving the computational efficiency required for real-time traffic forecasting.
Specifically, the encoder includes two attention modules:
a Spatial Attention Block, built upon our proposed DDC module, which flexibly captures spatial heterogeneity and non-Euclidean structures; subsequently, a Spatio-Temporal Attention Block that extends Involution~\cite{li2021involution} to the spatio-temporal domain, enabling dynamic modeling of spatio-temporal heterogeneity.
The decoder is deliberately lightweight, using simple feedforward layers to reduce inference cost. 
By restricting complex operations to the encoder, DDCN maximizes representation power without compromising real-time feasibility.

Overall, DDCN effectively bridges the gap between accuracy and scalability, making it well-suited for deployment in large-scale, real-time traffic prediction scenarios.
Our main contributions are summarized as follows:
\begin{itemize}
    \item We propose DDCN, a novel CNN-based architecture that achieves both high accuracy and efficiency for spatio-temporal traffic prediction.
    \item To effectively capture spatio-temporal heterogeneity and non-Euclidean structures with CNN, we design a DDC module that integrates deformable and dynamic convolution layers.
    \item We adopt a Transformer-style CNN backbone with an attention-based encoder and lightweight decoder, further enhance prediction performance while maintaining efficiency, ensuring scalability.
    \item We conduct extensive experiments on four real-world datasets and demonstrate that DDCN consistently outperforms prior methods in both predictive performance and efficiency.
\end{itemize}

\section{Preliminaries}
In this section, we introduce the basic concepts relevant to our study, including the grid-based data representation and problem formulation.

\subsubsection{Definition 1  (Grid Representation)} 
\begin{figure}[tbh]
\centering
\includegraphics[width=0.7\columnwidth]{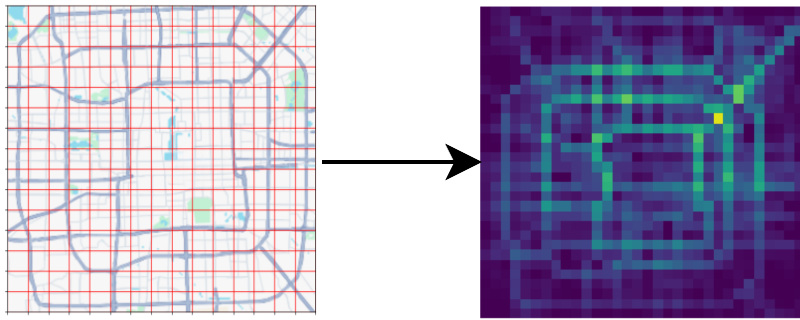}
\caption{Grid representation of traffic flow data.}
\label{fig1}
\end{figure}

As shown in Figure~\ref{fig1}, we partition the urban area into a grid map based on latitude and longitude coordinates.  
Specifically, the city is represented as a grid of size $H \times W$, where each cell corresponds to a fixed geographical region.
Each grid cell typically covers a fixed region of specific size of kilometers, providing a uniform spatial resolution.

\subsubsection{Definition 2 (Traffic Flow)} A spatio-temporal traffic flow map is defined as a five-dimensional tensor, $\textbf{X}_{1:T}=(\textbf{X}_1, \textbf{X}_2, ..., \textbf{X}_T) \in \mathbb{R}^{B \times T \times C \times H \times W}$. 
Here, $B$ and $C$ denote the batch size and the number of channels, representing the number of samples and the traffic flow features (e.g., inflow and outflow), respectively. 
$T$, $H$, and $W$ indicate the number of time steps, grid height, and grid width, respectively.
Each grid cell contains aggregated traffic flow data collected from sensors or records within the corresponding region.

\subsubsection{Problem Statement} Given a sequence of spatio-temporal traffic flow data $\textbf{X}_{1:T}$ as input, the goal is to predict the next traffic flow at $\textbf{X}_{T+1}$ time step.
Model $\mathcal{F}$ updates the weights $w$ as in equation (\ref{eq1}) to minimize the difference between $\textbf{X}_{T+1}$ and predicted value $\hat{\textbf{X}}_{T+1}$ using the loss function $\ell$.

\begin{equation}
\begin{split}
    &\hat{\mathbf{X}}_{T+1} = \mathcal{F}(\mathbf{X}_{1:T}) \\
    &w^{*} = \arg\min_{w} \ell(\hat{\mathbf{X}}_{T+1}, \mathbf{X}_{T+1})
\end{split}
\label{eq1}
\end{equation}

\section{Methodology}
In this section, we first introduce the Deformable Dynamic Convolution (DDC) module, which serves as a key component of our proposed model, DDCN.  
Then, we provide a detailed description of the overall architecture, including the backbone model and the rationale behind its design.

\begin{figure*}[t]
\centering
\includegraphics[width=1.0\textwidth]{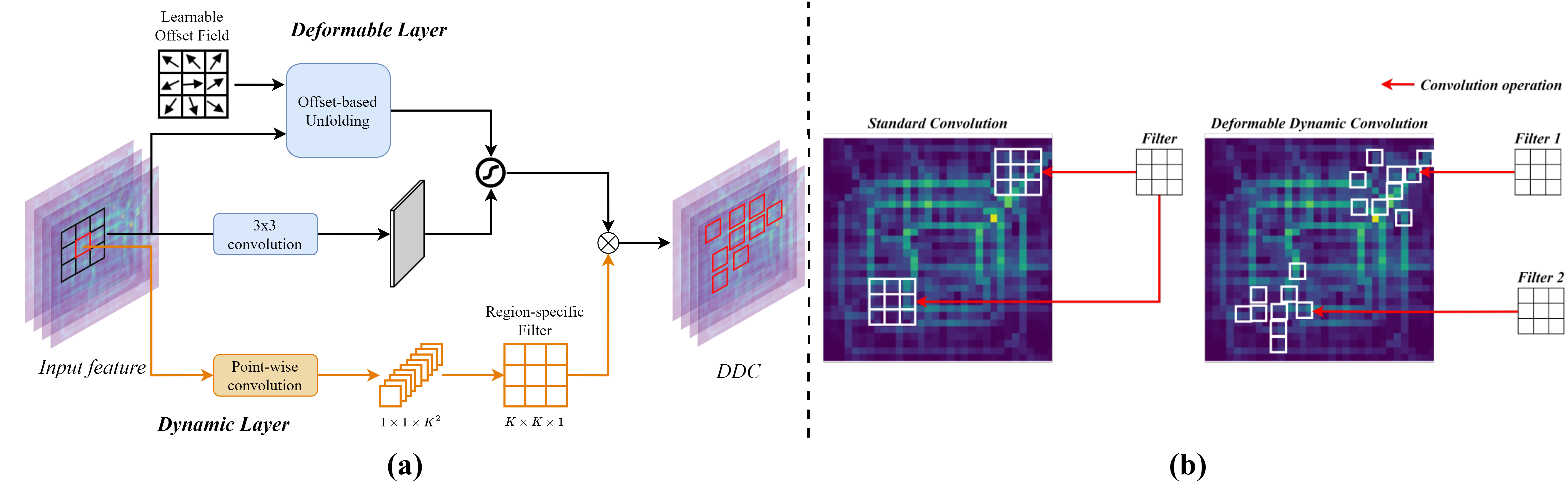}
\caption{(a): An illustration of proposed DDC module, (b): Comparison between standard and deformable dynamic convolution.}
\label{fig2}
\end{figure*}

\subsection{Deformable Dynamic Convolution}

As illustrated in Figure~\ref{fig2}, the proposed Deformable Dynamic Convolution (DDC) module is designed to simultaneously capture non-Euclidean spatial structures and spatial heterogeneity in traffic data.
The DDC module integrates two complementary components: a deformable layer and a dynamic layer.
The deformable layer enhances spatial flexibility by adjusting sampling positions based on learned offsets, while the dynamic layer generates region-specific filters conditioned on the input features.

\subsubsection{Deformable Layer}
In traditional convolution, features are sampled at fixed, grid-aligned positions.
To allow for spatial flexibility, the deformable layer predicts offsets $\Delta \mathbf{p}_k$ for each position in the receptive field, enabling the kernel to adapt to irregular spatial structures.
Formally, given an input feature map $\mathbf{X} \in \mathbb{R}^{C \times H \times W}$ and a convolution kernel with $K \times K$ sampling points, the output $\mathbf{Y}$ at position $\mathbf{p}_0$ is computed as:
\begin{equation}
    \mathbf{Y}(\mathbf{p}_0) = \sum_{k=1}^{K^2} w_k \cdot \mathbf{X}(\mathbf{p}_0 + \Delta \mathbf{p}_k)
\label{eq:deformable}
\end{equation}
where $w_k$ is the kernel weight at position $k$, and $\Delta \mathbf{p}_k$ is the learned offset for that position.
The offsets are predicted by a separate offset branch using a convolutional layer.

Then, to further refine the extracted local features, a 3×3 convolution followed by a sigmoid activation is applied to produce a spatial attention mask.
This mask is then element-wise multiplied with the unfolded feature, enabling the network to selectively emphasize or suppress specific regions within the deformable receptive field as follows:

\begin{equation}
\begin{split}
    \mathbf{A} &= \sigma(\text{Conv}_{3 \times 3}(\mathbf{X})) \\
    \mathbf{U} &=  f(y(\mathbf{p}_0))\\
    \tilde{\mathbf{U}} &= \mathbf{U} \cdot \mathbf{A} 
\end{split}
\label{eq:deformable2}
\end{equation}
Where $\sigma$, $\mathbf{A}$, $f$, $\mathbf{U}$, and $\tilde{\mathbf{U}}$ denotes sigmoid activation, spatial attention mask, unfold function, unfolded feature, and modulated feature map, respectively.

\subsubsection{Dynamic Layer}
While the deformable layer enhances spatial flexibility, it still shares the same kernel weights across all positions.
To address this limitation, we introduce a dynamic layer that generates region-specific filters conditioned on the input.
The structure of the dynamic layer is inspired by Involution~\cite{li2021involution}, which replaces traditional convolutional kernels with position-specific kernels that are dynamically generated from the input features at each spatial location.
This design allows the model to flexibly adapt to diverse spatial patterns and heterogeneity in traffic flow data.

Finally, the outputs of the deformable and dynamic layers are fused via element-wise multiplication of the sampled features and the dynamically generated weights.
This fusion enables both flexibility in spatial perception and adaptation to region-specific patterns.

Overall, the DDC module enriches spatial representations by adapting both where to look via deformable offsets, and how to extract features via dynamic kernels.
This dual adaptability is essential for modeling the complex and heterogeneous nature of traffic data in real-world urban environments.
In the following section, we detail how DDC is incorporated into the network and further extend its capability with spatio-temporal attention mechanisms.

\subsection{DDCN Structure}
\begin{figure*}[t]
\centering
\includegraphics[width=1.0\linewidth]{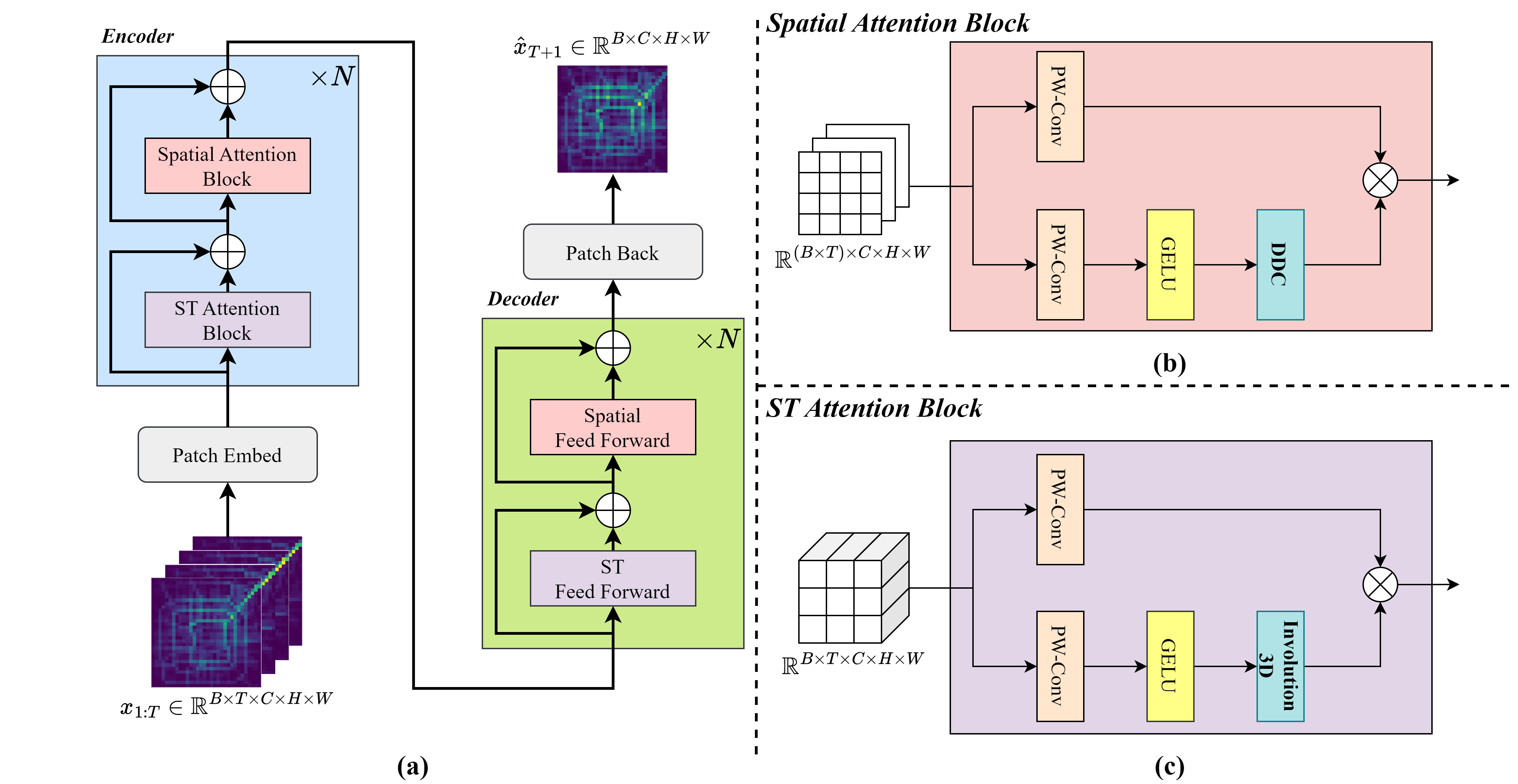}
\caption{(a): Overall structure of DDCN, (b): Detail structure of Spatial Attention Block, (c): Detail structure of Spatio-temporal (ST) Attention Block. $\oplus$ denotes element-wise add and $\otimes$ denotes for hadamard product.
The encoder captures spatio-temporal and spatial information through a ST Attention Block and a Spatial Attention Block, and passes them to the decoder. The decoder performs the role of combining the received information.}
\label{fig4}
\end{figure*}

The overall architecture of DDCN is shown in Figure~\ref{fig4} (a). 
The model follows an encoder–decoder framework, which is widely adopted in both sequence modeling and computer vision due to its flexibility in hierarchical representation learning. 
We are particularly inspired by recent advances in transformer-style convolutional architectures~\cite{jin2025datc}, which decouple spatial–temporal modeling from prediction, allowing deeper and more scalable networks.

To handle spatio-temporal grid data, the input sequence $\boldsymbol{X}_{1:T} \in \mathbb{R}^{B \times T \times C \times H \times W}$ is first passed through a patch embedding module that divides the spatial map into non-overlapping patches while increasing the number of channels:
\begin{equation}
    \boldsymbol{X}_{ST} = \text{PatchEmbed}(\boldsymbol{X}_{1:T}).
\end{equation}

\subsubsection{Encoder}  
The encoder consists of two types of attention blocks: the Spatio-Temporal Attention Block and the Spatial Attention Block, as illustrated in Figure~\ref{fig4}(b) and (c), respectively. These modules are stacked $N$ times to extract hierarchical representations.
Furthermore, the ST Attention Block extends Involution to spatio-temporal dimension, Involution$_{3D}$.
It allows dynamic kernel generation across both spatial and temporal dimensions.
For each center position $\mathbf{p}_0 = (t, h, w)$ in the input volume, a position-specific kernel $w_{\mathbf{p}_0}^{(k)}$ is dynamically generated and applied over the local neighborhood.
The output $\mathbf{Y}$ of Involution$_{3D}$ is calculated as follows:
\begin{equation}
\mathbf{Y}(\mathbf{p}_0) = \sum_{k=1}^{K^3} w_{\mathbf{p}_0}^{(k)} \cdot \mathbf{X}(\mathbf{p}_0)
\label{eq:involution3d}
\end{equation}

Overall, the ST Attention Block process is formulated as:
\begin{equation}
\begin{split}
    \boldsymbol{V}_{ST} &= \text{PW-Conv}_{3D}(\boldsymbol{X}_{ST}) \\
    \boldsymbol{Att}_{ST} &= \text{Involution}_{3D}(\sigma(\text{PW-Conv}_{3D}(\boldsymbol{X}_{ST}))) \\
    \text{ST Att Block} &= \boldsymbol{V}_{ST} \otimes \boldsymbol{Att}_{ST}
\end{split}
\label{eq4}
\end{equation}
Here, $\sigma$ denotes the GELU activation function. Involution$_{3D}$ generates content-dependent, region-specific kernels to enhance spatial and temporal heterogeneity modeling.

The output of the ST Attention Block is combined with the input via skip connection, and passed to the Spatial Attention Block. To isolate spatial features, the input is reshaped by merging the batch and time dimensions:
\begin{equation}
\begin{split}
    \boldsymbol{X}_{S} &= \text{ST Att Block} \oplus \boldsymbol{X}_{ST} \\
    \boldsymbol{V}_{S} &= \text{PW-Conv}(\boldsymbol{X}_{S}) \\
    \boldsymbol{Att}_{S} &= \text{DDC}(\sigma(\text{PW-Conv}(\boldsymbol{X}_{S}))) \\
    \text{Spatial Att Block} &= \boldsymbol{V}_S \otimes \boldsymbol{Att}_S
\end{split}
\label{eq5}
\end{equation}
The Spatial Attention Block applies our proposed Deformable Dynamic Convolution (DDC) to capture non-Euclidean structures and spatial heterogeneity. The learned attention is applied element-wise to the input features.

Each attention block thus computes dynamic region-specific attention maps and selectively emphasizes informative patterns. The output of the encoder is the result of stacking these modules with residual connections.

\subsubsection{Decoder}
The decoder aggregates the encoded features and performs the final prediction. It consists of a Feed Forward module composed of two point-wise convolution layers (PW-Conv), acting as a lightweight multi-layer perceptron. The decoder process is defined as:
\begin{equation}
\begin{split}
    Enc_{out} &= \text{Spatial Att Block} \oplus \boldsymbol{X}_{S} \\
    \text{FeedForward} &= \text{PW-Conv}(\text{PW-Conv}(Enc_{out})) \\
    Dec_{out} &= \text{FeedForward} \oplus Enc_{out} \\
    \boldsymbol{\hat{X}}_{T+1} &= \text{PatchBack}(Dec_{out})
\end{split}
\label{eq6}
\end{equation}
Here, the PatchBack module restores the spatial resolution and channel dimensions to match the target output format. By confining attention mechanisms to the encoder and using a simple decoder, DDCN reduces inference cost while maintaining strong predictive power.

\section{Experiments}
\subsection{Experimental Settings}
\subsubsection{Datasets}
We evaluate our model on four real-world spatio-temporal traffic datasets collected from New York City and Beijing. 
NYCBike1~\cite{zhang2017deep} and NYCBike2~\cite{yao2019revisiting} contain bike flow data from NYC, while NYCTaxi~\cite{yao2019revisiting} and BJTaxi~\cite{zhang2017deep} include taxi flow data from NYC and Beijing, respectively.
Table~\ref{table1} summarizes the key statistics of each dataset.

\begin{table}[t]
\centering
\caption{Detailed statistics of the traffic prediction datasets.}
\resizebox{1.0\linewidth}{!}{
\begin{tabular}{c c c c c}
    \hline
    Dataset & NYCBike1 & NYCBike2 & NYCTaxi & BJTaxi \\
    \hline
    Year & 2014 & 2016 & 2015 & 2015 \\
    Duration & 04/01–09/30 & 07/01–08/29 & 01/01–03/01 & 03/01–06/30 \\
    Interval & 1 hour & 30 min & 30 min & 30 min \\
    Grid Size & 16 $\times$ 8 & 10 $\times$ 20 & 10 $\times$ 20 & 32 $\times$ 32 \\
    Traffic Flows & 6.8K & 2.6M & 22M & 34K \\
    \hline
\end{tabular}
}
\label{table1}
\end{table}

During preprocessing, Min-Max normalization is applied to scale traffic flow values to the range [0, 1]. 
For evaluation, predicted values are inverse-transformed to their original scale.
We adopt a sliding window approach where the input consists of four consecutive time steps to predict the traffic flow at the next time step.
Each dataset is split into training, validation, and test sets in a 7:1:2 ratio.

\subsubsection{Evaluation Metrics}
We evaluate model performance using three common regression metrics: Root Mean Squared Error (RMSE), Mean Absolute Error (MAE), and Mean Absolute Percentage Error (MAPE).

RMSE calculates the square root of the average of squared differences between predicted and actual values, and is more sensitive to large errors due to the squaring operation.
MAE measures the average of the absolute differences between predicted value $\hat{y}$ and actual value $y$, providing a scale-dependent error.
MAPE calculates the average of the absolute percentage differences between predicted and actual values, offering an intuitive representation of prediction errors as a percentage of the ground truth.
Lower values of these metrics indicate more accurate predictions.
Each evaluation metric are represented as:

\begin{equation}
\begin{split}
    \text{RMSE} &= \sqrt{ \frac{1}{n} \sum_{i=1}^{n} (\hat{y}_i - y_i)^2 } \\
    \text{MAE}  &= \frac{1}{n} \sum_{i=1}^{n} \left| \hat{y}_i - y_i \right| \\
    \text{MAPE} &= \frac{100}{n} \sum_{i=1}^{n} \left| \frac{\hat{y}_i - y_i}{y_i} \right|
\end{split}
\label{eq:metric}
\end{equation}

\subsubsection{Baselines}
To validate the performance of DDCN, we compare it against a diverse set of strong baselines grouped into three categories:

\begin{itemize}
    \item \textbf{Traffic prediction methods:} 
    ST-ResNet~\cite{zhang2017deep} is a CNN-based model. 
    STGCN~\cite{yu2018spatio} is a GNN-based spatio-temporal model. 
    ST-SSL~\cite{ji2023spatio} incorporates self-supervised learning to better handle heterogeneity. 
    AdpSTGCN~\cite{chen2024adpstgcn} leverages adaptive graphs generated via multi-head attention.

    \item \textbf{Video prediction methods:} 
    ConvLSTM~\cite{shi2015convolutional} is a pioneering RNN-based video prediction model. 
    SwinLSTM~\cite{tang2023swinlstm} uses Swin Transformer-based attention.
    SimVPv2~\cite{tan2025simvpv2} is a CNN-only model optimized for spatio-temporal prediction efficiency.

    \item \textbf{Time-series forecasting methods:}
    PatchTST~\cite{Yuqietal-2023-PatchTST} is a Transformer-based model designed for long sequence forecasting. 
    iTransformer~\cite{liuitransformer} uses inverted architecture and variable embedding for temporal modeling.
\end{itemize}

\subsubsection{Implementation Details}
All experiments are implemented using Python 3.10.8 and PyTorch 2.1.1.
The models are trained on a machine running Ubuntu 20.04 with an Intel Core i7-10700 CPU and an NVIDIA RTX 3070 GPU.
We use the AdamW optimizer and train each model for 100 epochs with a batch size of 16.
For our proposed DDCN, the embedding dimension is set to 64, and the learning rate is set to 0.005.
The L1 loss function is used as the objective function during training.

\subsection{Experimental Results}

\setlength{\dashlinegap}{1.0pt}
\begin{table*}[t]
\centering
\caption{Quantitative results on the traffic prediction. \textbf{Bold} and \underline{\text{underline}} represent best and second performance, respectively.}
\resizebox{1.0\textwidth}{!}{
\begin{tabular}{c|c|c|c|c|c|c|c|c|c|c|c|c}
    \hline
    \multirow{2}{*}{Method} & \multicolumn{3}{c|}{NYCBike1} & \multicolumn{3}{c|}{NYCBike2} & \multicolumn{3}{c|}{NYCTaxi} & \multicolumn{3}{c}{BJTaxi} \\
    \cline{2-13}
    
    & \multicolumn{1}{c|}{RMSE} & \multicolumn{1}{c|}{MAE} & \multicolumn{1}{c|}{MAPE} & \multicolumn{1}{c|}{RMSE} & \multicolumn{1}{c|}{MAE} & \multicolumn{1}{c|}{MAPE} & \multicolumn{1}{c|}{RMSE} & \multicolumn{1}{c|}{MAE} & \multicolumn{1}{c|}{MAPE} & \multicolumn{1}{c|}{RMSE} & \multicolumn{1}{c|}{MAE} & \multicolumn{1}{c}{MAPE} \\
    \hline    
    \hline
    
    ST-ResNet~\cite{zhang2017deep} & 8.46 & 6.01 & 28.77 & 7.01 & 5.12 & 29.71 & \textbf{18.91} & 11.60 & 21.93 & 18.77 & 12.21 & 16.84 \\
    STGCN~\cite{yu2018spatio} & 8.01 & 5.69 & 26.81 & 7.43 & 5.34 & 30.63 & 23.01 & 13.72 & 22.73 & 19.12 & 12.29 & 16.53 \\
    ST-SSL~\cite{ji2023spatio} & 7.96 & 5.66 & 26.75 & 6.96 & 5.01 & 29.15 & 21.80 & 12.70 & 21.15 & 18.81 & 11.95 & \textbf{15.63} \\
    AdpSTGCN~\cite{chen2024adpstgcn} & \textbf{7.74} & \textbf{5.48} & 26.23 & \textbf{6.65} & \underline{4.83} & 27.81 & 19.76 & \underline{11.51} & 19.44 & \underline{18.40} & \underline{11.84} & 16.01 \\
    \hdashline
    
    ConvLSTM~\cite{shi2015convolutional} & 24.31 & 16.23 & 61.36 & 17.16 & 11.20 & 56.13 & 72.64 & 38.69 & 48.46 & 55.73 & 35.12 & 43.22 \\
    SwinLSTM~\cite{tang2023swinlstm} & 13.95 & 9.05 & 36.49 & 8.79 & 5.76 & 27.98 & 20.13 & 11.77 & \underline{19.36} & 21.03 & 13.43 & 17.62 \\
    SimVPv2~\cite{tan2025simvpv2} & \underline{7.95} & 5.61 & \underline{26.11} & 6.99 & 4.98 & 28.37 & 21.42 & 12.05 & 19.76 & 18.41 & 11.89 & 16.27 \\
    \hdashline
    
    PatchTST~\cite{Yuqietal-2023-PatchTST} & 13.21 & 8.84 & 36.89 & 9.77 & 6.58 & 33.79 & 28.82 & 16.25 & 23.75 & 25.11 & 16.01 & 20.45 \\
    iTransformer~\cite{liuitransformer} & 10.53 & 6.97 & 29.21 & 7.97 & 5.21 & \underline{25.45} & 25.85 & 14.67 & 20.29 & 22.53 & 14.31 & 18.53 \\
    \hline
    \textbf{DDCN (Ours)} & 8.09 & \underline{5.59} & \textbf{24.37} & \underline{6.87} & \textbf{4.71} & \textbf{24.72} & \underline{19.42} & \textbf{11.34} & \textbf{17.94} & \textbf{18.19} & \textbf{11.74} & \underline{15.74} \\
    \hline
\end{tabular}
}
\label{table2}
\end{table*}

\begin{table}[t]
\centering
\caption{Efficiency comparison results on spatio-temporal prediction models with BJTaxi dataset.}
\resizebox{1.0\linewidth}{!}{
\begin{tabular}{c c c c c}
    \hline
    Method & Params(M) & FLOPs(G) & RMSE & Train time(sec)\\
    \hline
    ST-ResNet & 1.19 & 1.22 & 18.77 & 3.55\\
    STGCN & 0.69 & 0.25 & 19.12 & 8.75\\
    ST-SSL & 1.26 & 6.05 & 18.81 & 40.62\\
    AdpSTGCN & 1.08 & 26.18 & 18.40 & 46.17\\
    \hdashline
    ConvLSTM & 3.97 & 12.23 & 55.73 & 25.14\\
    SwinLSTM & 2.77 & 2.64 & 21.03 & 26.97\\
    SimVPv2 & 10.80 & 0.73 & 18.41 & 9.62\\
    \hline
    \textbf{DDCN(Ours)} & \textbf{0.61} & \textbf{0.15} & \textbf{18.19} & \textbf{6.45}\\
    \hline
\end{tabular}
}
\label{table3}
\end{table}

\subsubsection{Quantitative Comparison Results}

Table~\ref{table2} and Table~\ref{table3} report the quantitative comparison and efficiency analysis results, respectively.
Compared to traffic prediction methods, DDCN demonstrates outstanding performance across all metrics.
While AdpSTGCN achieves the lowest RMSE on NYCBike1, DDCN consistently outperforms other methods in MAE and MAPE, particularly showing substantial improvements on NYCBike2 and NYCTaxi.
Notably, DDCN surpasses all state-of-the-art GNN-based methods in most cases, despite not relying on explicit graph structures or adjacency matrices.

This efficiency is further supported by Table~\ref{table3}, where DDCN achieves the lowest parameter count (0.61M) and FLOPs (0.15G), confirming its suitability for real-world deployment.
Although the training time of DDCN is slightly longer than that of ST-ResNet, this is primarily due to the additional overhead introduced by the deformable and dynamic computations, which involve learning input-adaptive sampling locations and region-specific kernels.

However, these dynamic operations significantly enhance the model's representational capacity, enabling DDCN to better capture complex and heterogeneous spatio-temporal patterns. As a result, the performance gain from the proposed DDC module outweighs the computational overhead, as evidenced by superior accuracy metrics across all datasets.
Moreover, DDCN still trains faster than attention-based baselines like AdpSTGCN and ST-SSL, demonstrating a favorable balance between prediction quality and computational efficiency. Note that the training time reported in Table~\ref{table3} corresponds to the average duration per epoch, providing a fair basis for efficiency comparison across models.

When compared with computer vision approaches, DDCN outperforms even highly efficient and accurate models like SimVPv2. These results highlight the ability of DDCN to effectively capture spatial heterogeneity and non-Euclidean patterns via its deformable dynamic convolution, unlike vision models that rely on standard CNN kernels. Moreover, the computational advantages of DDCN are evident, with significantly fewer parameters and FLOPs than models like ConvLSTM and SwinLSTM, confirming its efficiency.

For time-series forecasting models, such as PatchTST and iTransformer, performance is generally weaker. This reinforces the need for explicitly modeling spatio-temporal correlations in traffic data, as done in DDCN. While transformers offer strong temporal modeling capabilities, omitting spatial dependencies leads to suboptimal performance in traffic forecasting tasks.

Overall, DDCN achieves both state-of-the-art prediction accuracy and unmatched computational efficiency, confirming the effectiveness of integrating deformable dynamic convolution within a transformer-style encoder-decoder framework for large-scale spatio-temporal forecasting.

\subsubsection{Qualitative Comparison Results}
\begin{figure}[t]
\centering
\includegraphics[width=1.0\linewidth]{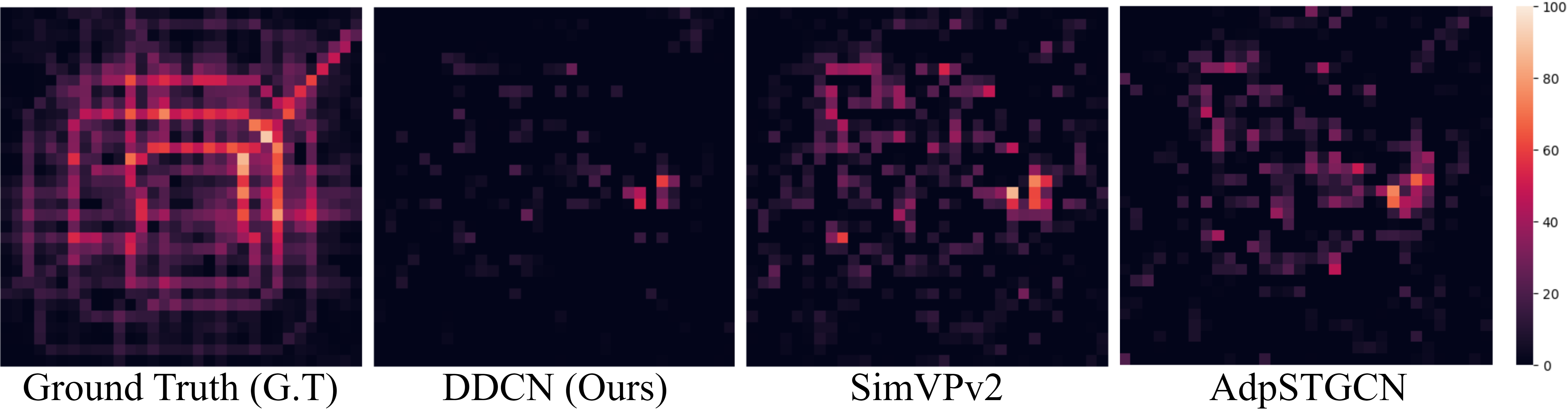}
\caption{An illustration of ground-truth and error maps per comparative models.}
\label{fig5}
\end{figure}

Figure~\ref{fig5} visualizes the ground-truth and corresponding error maps on the BJTaxi test set. The error maps are calculated as $\left | \hat{y} - y \right |$, where brighter areas indicate larger prediction errors.

As illustrated, DDCN accurately reconstructs key traffic regions, especially along major roads and intersections, with minimal residuals. This indicates the model's strong ability to capture both global traffic trends and localized variations. In contrast, SimVPv2 exhibits scattered and inconsistent error patterns, suggesting a lack of spatial coherence. While AdpSTGCN shows reasonable performance on densely trafficked areas, it suffers from higher errors in peripheral and less-structured regions.
These visual comparisons align well with the quantitative results presented earlier, reinforcing the effectiveness of the proposed deformable dynamic convolution. By adaptively adjusting receptive fields and kernel responses, DDCN successfully focuses on spatially heterogeneous regions and models non-Euclidean patterns that are often overlooked by conventional CNN- or GNN-based approaches.

\begin{table}[t]
\centering
\caption{Ablation study of DDCN.}
\resizebox{0.7\linewidth}{!}{
\begin{tabular}{c c c c}
    \hline
    Method & RMSE & MAE & MAPE \\
    \hline
    w/o. all & 19.92 & 12.36 & 16.45 \\
    w/o. Involution$_{3D}$ & 19.12 & 11.99 & 16.44 \\
    w/o. DDC & 18.62 & 11.97 & 16.14 \\
    \hline
    \textbf{DDCN(Ours)} & \textbf{18.19} & \textbf{11.74} & \textbf{15.74} \\
    \hline
\end{tabular}
}
\label{table4}
\end{table}

\subsection{Ablation Study}
To evaluate the contribution of key components in DDCN, we conduct an ablation study using the BJTaxi dataset. Results are summarized in Table~\ref{table4}.

Removing the deformable dynamic convolution (w/o. DDC) significantly impairs spatial modeling, confirming its role in adapting to irregular patterns. Excluding the spatio-temporal Involution$_{3D}$ (w/o. Involution$_{3D}$) also results in performance degradation, indicating the importance of capturing temporal heterogeneity. When both modules are removed (w/o. all), prediction accuracy declines the most, underscoring the complementary benefits of both modules.

Overall, these results confirm that both the DDC module and Involution$_{3D}$ play vital roles in improving spatio-temporal prediction accuracy.

\section{Related Work}

\subsection{Traffic Prediction}
Traffic prediction aims to forecast future traffic conditions by learning from past spatio-temporal patterns. Early methods based on statistical models~\cite{smith1997traffic,yu2004switching,caliendo2007crash} and traditional machine learning techniques~\cite{lv2009real,sharma2016traffic} struggled to capture the complex interactions between space and time, as they mainly focused on temporal sequences~\cite{luo2024lsttn,ahmed2024enhancement}.

To address this, deep learning models that jointly consider spatial and temporal dependencies have emerged. LCGBN~\cite{zhu2016short} and ST-ResNet~\cite{zhang2017deep} highlighted the importance of integrating both dimensions using probabilistic and CNN-based approaches, respectively.
With the recognition of traffic data's non-Euclidean structure, GNN-based methods have become popular. STGCN~\cite{yu2018spatio} introduced a hybrid of graph and temporal convolutions. GMAN~\cite{zheng2020gman} further incorporated attention mechanisms within a GNN encoder–decoder. ST-SSL~\cite{ji2023spatio} adopted a self-supervised framework with adaptive graph augmentation, and AdpSTGCN~\cite{chen2024adpstgcn} proposed generating graph structures dynamically via attention.

Despite their effectiveness, GNN-based models rely on adjacency matrices and incur high computational costs~\cite{ma2022graph,duan2022comprehensive}, which can hinder scalability in large-scale traffic forecasting~\cite{jyothi2023graph}.

\subsection{Variant Convolutions}
Standard convolutions apply fixed filters across the spatial domain, which limits their ability to capture spatial heterogeneity, a key aspect of traffic data.

To address this, deformable convolutions~\cite{dai2017deformable} introduced learnable offsets, allowing adaptive receptive fields. Kervolution~\cite{wang2019kervolutional} extended convolutions into non-linear kernel space, and Involution~\cite{li2021involution} proposed region-specific dynamic kernels that have proven effective in various tasks~\cite{ye2022dense,zhu2025towards}.
And recently, Kim et al.~\cite{kim2023deformable} further combined deformable and dynamic kernels into a unified framework, enhancing spatial flexibility and adaptability.

These advances motivate our work, which integrates deformable and dynamic convolutions into a CNN-based architecture tailored for traffic prediction.
Our approach captures spatial irregularities without relying on graph structures, achieving high accuracy and efficiency simultaneously.

\section{Discussion}
While DDCN demonstrates strong prediction accuracy and computational efficiency across diverse real-world datasets, several challenges and opportunities remain for further enhancement.
First, although DDCN effectively captures spatial heterogeneity and non-Euclidean structures through its deformable dynamic convolution, it still operates on a grid layout. 
This spatial discretization may limit adaptability in urban areas with irregular road topologies or sparse infrastructure.
Future work could explore hybrid spatial representations that combine the flexibility of graph-based modeling with the regularity and efficiency of grid structures, thereby enhancing spatial expressiveness.

Second, despite introducing some computational overhead from learning input-adaptive offsets and region-specific kernels, the deformable dynamic convolution maintains a favorable efficiency–accuracy trade-off. 
As evidenced by our results, DDCN achieves superior accuracy with significantly fewer parameters and FLOPs compared to other baselines.
In future work, we aim to further optimize the DDC module to reduce training overhead while preserving its representational strength.

Third, the current model assumes uniform time intervals and a fixed historical window size. 
However, real-world traffic data often contains irregular temporal patterns, missing values, or sensor outages.
Incorporating temporal attention mechanisms or time-aware masking strategies could improve robustness under such conditions.

In addition, while DDCN outperforms all baselines in RMSE and MAE, it exhibits slightly higher MAPE than ST-SSL on the BJTaxi dataset. 
As noted in~\cite{hyndman2006another}, MAPE tends to inflate percentage errors in low-value regions where even small absolute deviations can result in disproportionately large percentage errors.
This observation highlights the need for cautious interpretation of MAPE, especially in sparse traffic zones.

For future directions, we plan to evaluate DDCN on more fine-grained grid maps that include minor or missing road segments to further assess spatial generalization. 
We also aim to incorporate external contextual variables—such as weather, holidays, and events—to enrich temporal understanding.
Lastly, cross-city generalization via transfer learning and domain adaptation will be explored to enhance the scalability of DDCN in diverse urban environments.
In summary, DDCN presents a promising direction for traffic forecasting by bridging CNN-based modeling with spatial adaptivity and dynamic learning. 
Its efficiency and flexibility make it well-suited for real-world intelligent transportation systems.

\section{Conclusion}
In this paper, we introduced Deformable Dynamic Convolutional Network (DDCN), a novel CNN-based architecture designed for efficient and accurate spatio-temporal traffic prediction. 
By incorporating deformable and dynamic convolutions within a transformer-style encoder–decoder framework, DDCN flexibly models spatial heterogeneity and non-Euclidean patterns without relying on explicit graph structures.
Extensive experiments on four real-world datasets demonstrated that DDCN achieves state-of-the-art or competitive results across multiple evaluation metrics, while significantly reducing model complexity in terms of parameters and FLOPs.
Ablation studies further confirmed the complementary roles of deformable and dynamic modules in boosting performance.

The results highlight that CNN-based architectures, when enhanced with spatial adaptivity and dynamic filtering, can serve as a viable alternative to GNN-based methods for large-scale traffic forecasting.
Our approach opens new directions for developing lightweight, scalable, and generalizable models tailored to practical intelligent transportation applications.

\bibliographystyle{IEEEtran}
\bibliography{ref}

\end{document}